\pdfoutput=1

\documentclass[11pt]{article}

\usepackage{acl}

\usepackage{times}
\usepackage{latexsym}

\usepackage[T1]{fontenc}

\usepackage[utf8]{inputenc}

\usepackage{microtype}

\usepackage{graphicx}
\usepackage{booktabs}
\usepackage{multirow}

\title{Transformer-based HTR for Historical Documents}

\author{Phillip Benjamin Ströbel \and Simon Clematide \and Martin Volk \\
  Department of Computational Linguistics\\
  University of Zurich\\
  \texttt{\{pstroebel,siclemat,volk\}@cl.uzh.ch} \\\AND
  Tobias Hodel \\
  Walter Benjamin Kolleg \\
  University of Bern\\
  \texttt{tobias.hodel@unibe.ch} \\}

\begin{document}
\maketitle

\section{Introduction}
\textit{Handwritten Text Recognition} (HTR) has become a valuable tool to extract text from scanned documents \citep{terras2021inviting}. The current digitisation wave in libraries and archives does not stop at historical manuscripts. As such, HTR plays an essential role in making the contents of manuscripts available to researchers and the public. 

HTR has undergone significant improvements in recent years, thanks in large part to the introduction of neural network-based techniques \citep{graves2008offline,graves2009novel}. Platforms like \textit{Transkribus}\footnote{\url{https://readcoop.eu/de/transkribus/}} successfully integrated these approaches in a way that its HTR+ model \citep{michael2018htr} can achieve character error rates (CERs) of below 5\% with little annotated ground truth material \citep{Mueh19}.

\begin{figure}
    \centering
    \includegraphics[width=\columnwidth]{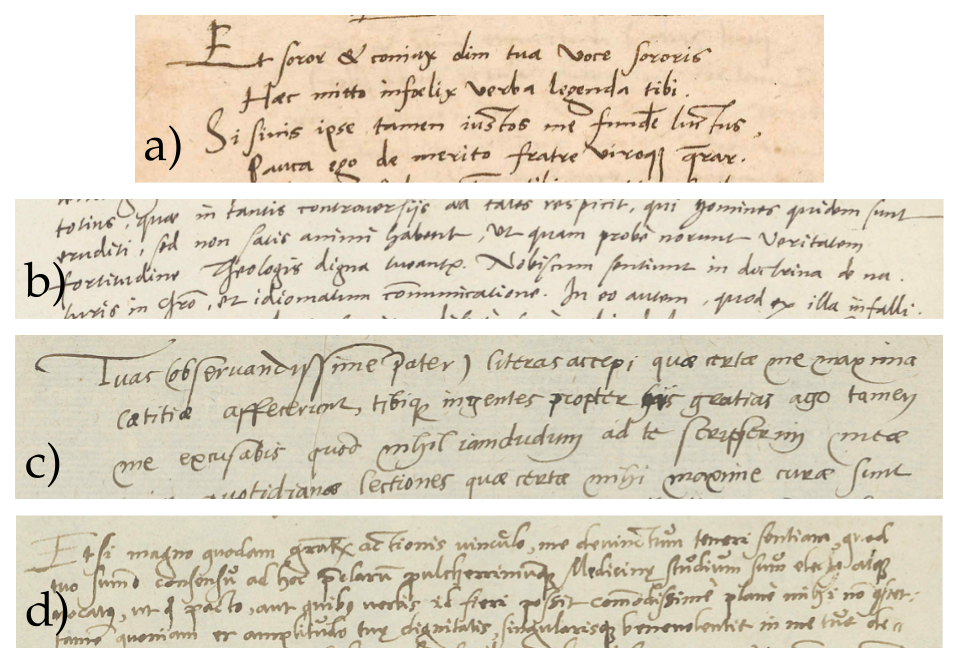}
    \caption{Different handwriting styles. Poetry by Rudolf Gwalther (a), letters by  Rudolf Gwalther (b), Matthieu Coignet (b), and  Kaspar Wolf (c).}
    \label{fig:hwstyles}
\end{figure}

However, a look at the digital platform for manuscript material for Swiss libraries and archives  \textit{e-manuscripta}\footnote{\url{https://www.e-manuscripta.ch/}} shows that in the category ``correspondence'' containing 45k titles, only 313, or $0.1\%$, contain transcriptions. Such large manuscript collections pose significant challenges to libraries and archives, especially because of the variety of handwriting styles. That the authors' handwriting changes according to what they were writing only adds in complexity. Fig.~\ref{fig:hwstyles} exemplifies this by showing Rudolf Gwalther's handwriting in (a) a 16\textsuperscript{th} century poetry volume and (b) a letter, among other handwritings from different authors (c and d).

The variability of such collections calls for models that adapt well to different hands with only little to no training data. Transformer-based architectures \citep{vaswani2017attention} have proven suitable to build large language representation models like, e.g., BERT \citep{devlin2018bert}. BERT-style models are used to fine-tune specific models for natural language understanding and are known as strong transfer learners \citep{ruder2019transfer}. Most recently, transformers have found their way into image processing \citep{dosovitskiy2020image,touvron2021training}, which drove the development of image transformers \citep{bao2021beit}.

\section{Approach}
The basis for our research is TrOCR \cite{li2021trocr}, which combines  the BERT-style vision transformer BEiT \citep{bao2021beit} with a RoBERTa \citep{liu2019roberta} language representation model. BEiT works as an encoder and is pre-trained on the Image-Net-1K \citep{russakovsky2015imagenet} dataset containing 1.2M images, while RoBERTa serves as a decoder producing the text. \citet{li2021trocr} used 687M of printed and about 18M of synthetically generated handwritten text lines in English to pre-train the TrOCR model. During this phase, the model learns to extract relevant features from the images and decode them into English text, therefore training the language model from scratch. The authors initialised the RoBERTa decoder with 6 and 12 layers, referring to them as BASE when paired with the pre-trained 12 layer BEiT instance and LARGE when paired with the 24-layer BEiT model, respectively.

Finally, \citet{li2021trocr} fine-tuned their pre-trained TrOCR instances on ``real-world'' data, like the IAM dataset \citep{marti2002iam}. The IAM dataset consists of handwritten English lines from different authors.  TrOCR\textsubscript{BASE} reaches a CER of 3.42\% and TrOCR\textsubscript{LARGE} a CER of 2.89\% on this dataset. The score of TrOCR\textsubscript{LARGE} is only 0.14 percentage points behind the best score of \citet{diaz2021rethinking}, who used a different approach.

Our research aims to exploit the pre-trained vision and language transformers, hoping that a model fine-tuned on historical manuscripts generalises well enough to be applied to extensive and variable manuscript collections. We want to test whether we can transfer the ``knowledge'' about handwriting in the English language TrOCR has acquired early modern manuscripts.

\section{Data}
Our data stem from the 16\textsuperscript{th} century volume \textit{Lateinische Gedichte} by Rudolf Gwalther.\footnote{\url{https://doi.org/10.7891/e-manuscripta-26750}} \citet{stotz21gwalther} downloaded the available images and partial transcriptions from \textit{e-manuscripta} and loaded them into the Transkribus interface. They applied layout recognition to identify lines and baselines and aligned them with the transcriptions. The publicly available 
dataset has 4,037 image and corresponding text lines in Latin, which we split into 3,603 lines for training and 433 lines for validation.\footnote{\url{https://doi.org/10.5281/zenodo.4780947}}

A second dataset consists of 16,584 lines in Latin from Heinrich Bullinger's (1504 - 1575) correspondence. It contains hands from about 60 different authors with a heavily skewed author distribution. We split the data into 13,843 lines for training, 1,685 lines for validation, and 1,056 for testing.
%

\section{Experiments and Discussion}
We trained Transkribus HTR+ models on the Gwalther and Bullinger data for 50 epochs as reference models.\footnote{We used the \textit{Acta\_17 HTR+} as a base model.} Table \ref{tab:res} shows the result under ``HTR+''.

For the TrOCR architecture, using the same data, we fine-tuned both TrOCR\textsubscript{BASE} and TrOCR\textsubscript{LARGE} for three up to 20 epochs.\footnote{The untrained TrOCR\textsubscript{LARGE} model achieves a CER of 57.48\% on the validation data.}

Table \ref{tab:res} presents the results of our initial experiments: the longer we fine-tune the models, the better their performance gets. This effect is less pronounced for TrOCR\textsubscript{BASE}, however, where the performance even drops if we fine-tune more than ten epochs. Moreover, we note a clear performance gap between TrOCR\textsubscript{BASE} and TrOCR\textsubscript{LARGE}, where TrOCR\textsubscript{LARGE} always performs better. 

Our results are surprising because the pre-trained TrOCR model never saw any Latin data previous to our experiments. For example, our model only sees 23k Latin words during fine-tuning on the Gwalther data. The vocabulary overlap of the training and validation set is 68.9\%. Moreover, TrOCR has never been confronted with early modern manuscripts. Nevertheless, we achieved a CER that beats our reference model trained on Gwalther data at 0.19 percentage points on the validation set and 4.6 percentage points on the Bullinger data on the test set.

We, therefore, assume that TrOCR is a robust and highly transferable handwriting representation model that is suitable for being fine-tuned on hands of all styles and origins.

\begin{table}[]
\centering
\resizebox{\columnwidth}{!}{
\begin{tabular}{@{}ll|rrrrr|r@{}}
\toprule
                &                            & \multicolumn{5}{c|}{\textbf{fine-tuning epochs}}                                & \textbf{epochs} \\ 
\textbf{System} & \textbf{data}     & \textbf{3} & \textbf{5} & \textbf{10}   & \textbf{15}   & \textbf{20} & \textbf{50}     \\ \midrule
HTR+   & \multirow{3}{*}{Gwalther}  & -          & -          & -             & -             & -           & 2.74            \\
TrOCR\textsubscript{BASE}     &                            & 3.84       & 3.72       & \textbf{3.18} & 3.31          & 3.62        & -               \\
TrOCR\textsubscript{LARGE}    &                            & 2.94       & 2.72       & 2.58          & \textbf{2.55} & 2.62        & -               \\ \midrule
HTR+            & \multirow{2}{*}{Bullinger} & -          & - & -             & -             & -           & 21.13           \\
TrOCR\textsubscript{LARGE}    &                            & -          & -          & -    & 16.53         & -           & -               \\ \bottomrule 
\end{tabular}
}
\caption{CERs for different models and different (fine-tuning) epochs on the validation set for Gwalther data and the test set for Bullinger data.}
\label{tab:res}
\end{table}


\section{Conclusion}
Our initial experiments with TrOCR indicate that it outperforms state-of-the-art models for single-author and multi-author datasets. Astonishing is its strong performance on a language and handwriting styles it has never ``learnt to read''. Moreover, TrOCR does not require baseline information, in contrast to Transkribus models. 

In future experiments, we want to investigate whether plugging in a pre-trained Latin RoBERTa decoder plus adapting the encoder to early modern handwriting can improve performance.

Moreover, we want to further examine TrOCR on more variable datasets. For example, projects focusing on correspondences would benefit from  HTR models that adapt to many different authors. Thus, we will investigate whether TrOCR generalises better to this data than conventional methods.




\bibliography{anthology,comhum_2022_abstract}
\bibliographystyle{acl_natbib}

\end{document}